%% file: main.tex
\documentclass[runningheads]{llncs}

\usepackage{eccv}

\usepackage{eccvabbrv}

\usepackage{graphicx}
\usepackage{booktabs}
\usepackage{enumitem}
\usepackage{multicol}
\usepackage{multirow}
\usepackage{algorithm}
\usepackage{algorithmic}
\usepackage[accsupp]{axessibility}  

\usepackage[pagebackref,breaklinks,colorlinks,citecolor=eccvblue]{hyperref}

\usepackage{orcidlink}

\begin{document}

\title{
UniTranslator: A Unified Multi-modal Framework for End-to-end In-Image Machine Translation
} 

\titlerunning{Unitranslator}

\author{Jiahao Lyu\inst{1,2,5} \and
Pei Fu\inst{2}* \and 
Zhenhang Li\inst{4} \and
Shaojie Zhang\inst{2} \and
Jiahui Yang\inst{2} \and
Yu Zhou\inst{3}\textdagger \and
Can Ma\inst{1} \and
Zhenbo Luo\inst{2} \and
Jian Luan\inst{2}
}

\authorrunning{J.Lyu et al.}

\institute{Institute of Information Engineering, Chinese Academy of Science, China 
\\
\and
MiLM Plus, Xiaomi Inc, China
\\
\and
VCIP $\&$ TMCC $\&$ DISSec, College of Computer Science $\&$ College of Cryptology and Cyber Science, Nankai University, China
\and
Binghamton University, USA
\\
\and School of Cyber Security, University of Chinese Academy of Science, China 
}

\maketitle

\let\thefootnote\relax
\footnotetext{Work done when Jiahao Lyu was an intern at Xiaomi. *Project Leader. \textdagger Corresponding Author.} 

\vspace{-15pt}
\begin{abstract}
In-Image Machine Translation (IIMT) aims to translate scene text in an image and render the translated text back into the original regions while preserving the overall visual appearance. Recent unified multimodal models provide a promising solution by combining visual-text understanding and image generation within a single framework. However, directly adapting such models to IIMT remains challenging. In particular, they often suffer from understanding-generation conflicts, where the translation inferred during understanding is inconsistent with the text supervision used in generation, and spatial position misalignment, where the rendered text does not accurately match the target text regions.
To address these issues, we present UniTranslator, a unified multimodal framework for IIMT that tightly couples translation understanding and text editing. Specifically, we introduce an Understand-Generation Alignment Module (UGAM) to bridge the representation gap between understanding and generation, encouraging semantic consistency between translated content prediction and text rendering. We further propose a Spatial Mask Decoder (SMD) with pixel-level supervision over text regions to improve spatial grounding, geometric alignment, and layout controllability during generation.
Extensive experiments on multiple benchmarks demonstrate that UniTranslator achieves state-of-the-art performance across diverse language directions and complex real-world layouts. Moreover, our results reveal a strong mutual reinforcement effect between translation understanding and image generation, highlighting the advantage of unified translation multimodal learning. 
Code is available \href{https://github.com/SeerRay-Lab/Unitranslator}{here}.
\vspace{-10pt}
  \keywords{In-Image Machine Translation \and Unified Generation and Understanding Model \and Multi-modal Learning}
\end{abstract}

\vspace{-10pt}
\section{Introduction}
\label{sec:intro}

\begin{figure}[t]
\centering
\includegraphics[width=0.9\textwidth]{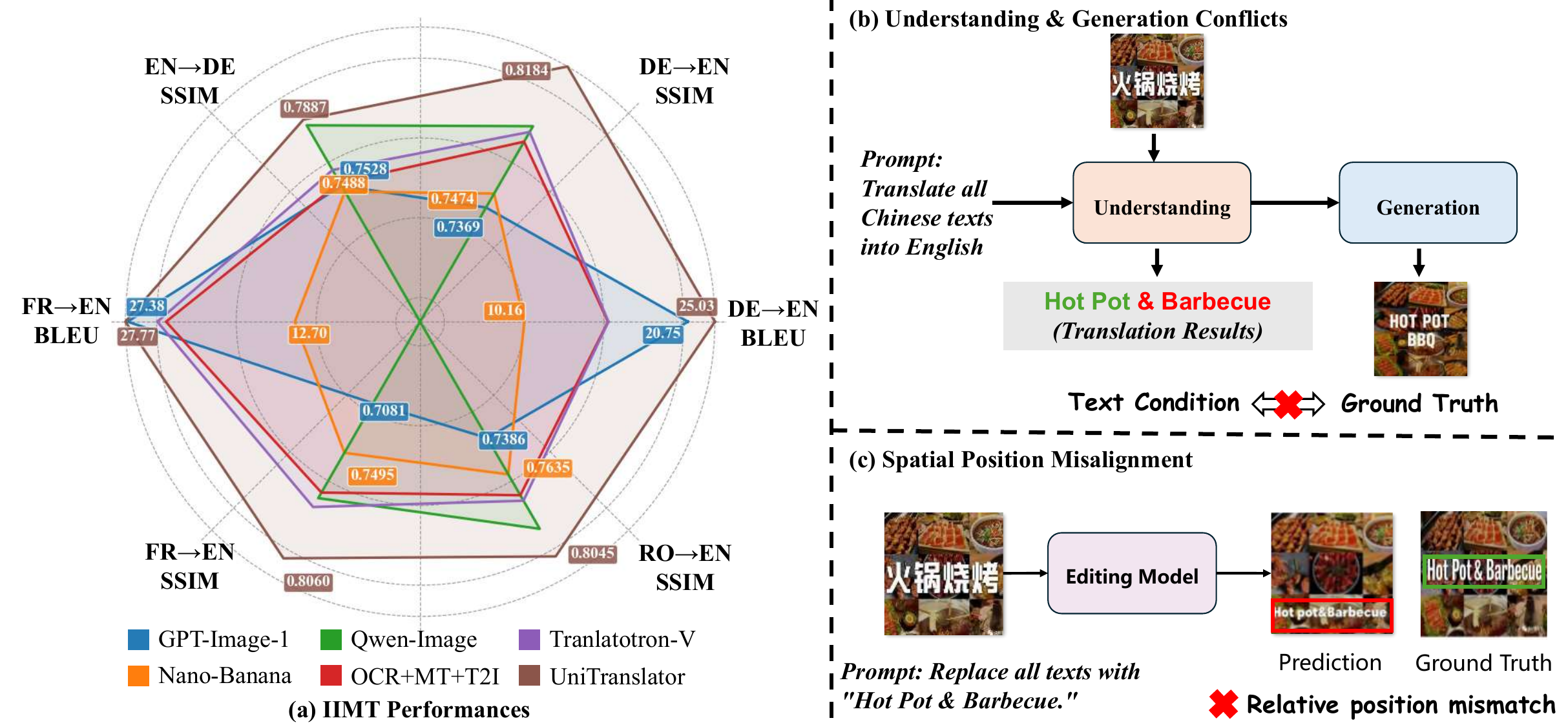}
    \caption{
    (a) Performance comparison of unified multimodal models and specialized image-to-image translation models on the Translatotron-V(ision) benchmark. (b) Illustration of the semantic conflict between the understanding and generation objectives in unified multimodal models for the IIMT task, which could cause the generated output to deviate from the target translation semantics. (c) Illustration of the positional inaccuracy issue in existing image editing models for text editing, where the edited text is often placed at an incorrect relative position.
}
    \label{fig:intro}
    \vspace{-20pt}
\end{figure}


\textbf{I}n-\textbf{I}mage \textbf{M}achine \textbf{T}ranslation (IIMT) aims to translate scene text in an image into a target language and write the translated content back into the original text regions while preserving the surrounding visual appearance. As a fundamental capability for multilingual visual communication, IIMT supports a broad range of applications, including travel assistance, augmented reality, and cross-border e-commerce, where translated text must be both semantically accurate and visually coherent with the scene. Most existing systems follow a cascaded OCR–MT–Editing pipeline~\cite{shu2025visual,zeng2024textctrl,li2024first,chen2026styletextgen}, which is often efficient but prone to error accumulation across stages.
Recent progress in unified multimodal models (UMMs) introduces a novel paradigm for IIMT. By integrating visual-text understanding and image generation within a single pre-trained framework, UMMs offer the potential to handle diverse languages, open-domain scenes, and complex layouts in a more generalizable manner than task-specific pipelines. Moreover, the strong performance of proprietary systems such as GPT-Image-1~\cite{gptimage1} suggests that this paradigm has a high capability ceiling for image-level text translation and editing. However, as shown in \cref{fig:intro}(a), existing UMMs adapted to IIMT, such as Nano-Banana~\cite{comanici2025gemini} and Qwen-Image~\cite{wu2025qwen}, still deliver unsatisfactory performance, even on relatively constrained benchmarks. This gap indicates that effectively unlocking the IIMT capability of pre-trained UMMs remains a meaningful and unresolved research problem.

A central obstacle lies in the mismatch between translation understanding and text generation. As illustrated in \cref{fig:intro}(b), the understanding branch first reads the source text and predicts a translation, which is then used to guide generation. However, a source expression can admit multiple valid translations. The translation predicted by the understanding branch may therefore be semantically correct but still differ from the exact target string shown in the ground truth image. In the example of \cref{fig:intro}(b), the model can produce a valid English translation such as ``\textit{Hot Pot $\&$ Barbecue}''. At the same time, the edited target image corresponds to a different rendering. As a result, the model receives inconsistent training signals: the understanding branch is encouraged to produce a valid translation, whereas the generation branch must reproduce a specific visual text instance. This discrepancy creates an \textbf{understanding-generation conflict} that hinders unified optimization for IIMT.

Beyond semantic consistency, IIMT also requires precise spatial control over where translated text is written back. As shown in \cref{fig:intro}(c), even when the target text content is largely correct, a generic unified image editing model can still place it in the wrong region or with an incorrect local arrangement. In the illustrated example, the model generates ``\textit{Hot Pot Barbecue}'', but fails to align it with the position and layout of the ground-truth text region. This reveals a key limitation of generic UMMs: they lack explicit supervision for fine-grained correspondence between source text regions and edited text placement. 
Diffusion models inherently favor global coherence during the denoising process, they are prone to altering the original background textures and compromising the rigid spatial placement of text regions.
Consequently, they struggle with the \textbf{spatial position misalignment} that is critical in IIMT, where semantic correctness and geometric fidelity are both required.

To address these challenges, we propose \textbf{UniTranslator}, a unified multimodal model for in-image machine translation that integrates translation understanding and visual text editing within a single framework. The core of UniTranslator is an \textbf{Understand-Generation Alignment Module (UGAM)}, which explicitly bridges the understanding and generation branches by aligning the translation-related representations used by the two modules. Rather than optimizing the two stages under disconnected objectives, UGAM encourages the model to maintain semantic consistency between translation understanding and visual text rendering, thereby alleviating the understanding-generation conflict at the representation level.
To further improve spatial precision, UniTranslator incorporates a \textbf{Spatial Mask Decoder (SMD)} that introduces pixel-level supervision over text editing regions. SMD injects explicit spatial cues into the generation process, enabling more accurate localization, geometric alignment, and layout-preserving text replacement. Together, UGAM and SMD address the two key bottlenecks of unified multimodal IIMT, namely semantic inconsistency between understanding and generation, and spatial position misalignment during text rendering.
Extensive experiments on multiple in-image translation benchmarks show that UniTranslator achieves state-of-the-art performance and generalizes effectively across diverse language directions and complex real-world scenes. In addition, we observe a clear mutual reinforcement effect between understanding and generation during training, indicating that accurate translation comprehension and faithful visual text editing can be learned synergistically within a unified multimodal framework.

Our main contributions are summarized as follows:

\begin{itemize}[leftmargin=*]
\item We present {UniTranslator}, a unified multimodal model for in-image machine translation that integrates translation understanding and visual text editing within a single framework.

\item We propose an {Understand-Generation Alignment Module} to mitigate the semantic conflict between understanding and generation, and a 
{Spatial Mask Decoder} to improve spatial accuracy in visual text editing through pixel-level supervision.

\item UniTranslator achieves state-of-the-art results on multiple in-image translation benchmarks, with strong generalization across diverse languages and complex real-world scenes. We further find that translation understanding and image generation reinforce each other during unified training.

\end{itemize}

\section{Related Works}

\subsection{In-Image Machine Translation}

Research on multimodal translation has mainly focused on \emph{text image translation} (TIT), where the goal is to translate text appearing in an image into target-language \emph{text}~\cite{chen2021cross,su2021rtnet,zhu2023peit,lan2023exploring,liang2024document,zhang2025reading,salesky2024benchmarking,zhang2025understand,wang2025rethinking,li2026mmtit}. In contrast, \emph{in-image machine translation} (IIMT) requires translating scene text and writing the translated content back into the image while preserving the original visual appearance, layout, and surrounding context.
Existing IIMT methods can be broadly divided into cascaded systems and end-to-end models. Cascaded approaches, such as AnyTrans~\cite{qian2024anytrans} and VisTrans~\cite{vaidya2024show}, decompose the task into multiple stages, including text detection, recognition, translation, and image editing. End-to-end methods instead directly model the full translation-and-editing process using synthetic supervision~\cite{tian2023image,lan2024translatotron} or subtitle-based data construction~\cite{tian2025exploring,tian2025prim,lyu2026imtbench}. Despite recent progress, existing methods still struggle to generalize across diverse languages, complex layouts, and real-world scenes. This motivates a more unified framework with stronger cross-lingual and cross-scenario generalization. Unified multimodal models (UMMs), which integrate visual understanding and image generation within a single pre-trained architecture, provide a promising basis for this goal.

\subsection{Unified Multi-modal Models}

Recent unified multimodal models (UMMs) aim to integrate visual understanding and image generation within a single framework, making them a natural foundation for IIMT. Existing UMMs can be broadly categorized into diffusion-based, auto-regressive, and hybrid architectures~\cite{zhang2025unified,jin2025efficient}.
Diffusion-based UMMs couple language and vision representations during denoising-based generation. Representative examples include Dual Diffusion, UniDisc~\cite{swerdlow2025unified}, FUDOKI~\cite{wang2025fudoki}, Muddit~\cite{shi2025muddit}, and MMaDA~\cite{yang2025mmada}. Auto-regressive models represent images as discrete token sequences and perform multimodal generation in a next-token prediction manner, including TokLIP~\cite{lin2025toklip}, Harmon~\cite{wu2025harmonizing}, Chameleon~\cite{team2024chameleon}, Emu3~\cite{wang2024emu3}, OmniGen~\cite{xiao2025omnigen}, UniWorld~\cite{lin2025uniworld}, and ILLUME~\cite{huang2025illume+}. Hybrid methods, such as Show-o~\cite{xie2024show} and BAGEL~\cite{deng2025emerging}, combine symbolic reasoning with continuous image generation to better balance controllability and visual fidelity.
Despite this progress, existing UMMs still fall short on IIMT, which requires tight coordination between text understanding and image editing, especially for content preservation and style-consistent text rendering. Our work addresses this gap by introducing explicit mechanisms for understand-generation alignment and spatially grounded text editing within a unified multimodal framework.
\section{Methods}

\subsection{Formulation}


Given a source image $I_{\mathrm{src}}$ containing text in a source language $s$ and a target language $t$, the goal of in-image machine translation (IIMT) is to produce both the translated target text $T_{\mathrm{tgt}}$ and a translated image $I_{\mathrm{tgt}}$, where the source text is replaced by its target-language translation while the non-text visual content and overall layout are preserved.
We formulate IIMT as a joint understanding-and-generation problem under a unified multimodal model $f_{\theta}$. Specifically, the model takes as input the source image $I_{\mathrm{src}}$ together with an instruction prompt $P(s,t)$, ``Translate all texts from $s$ to $t$.'' Conditioned on both the image and the instruction, the model jointly predicts the translated text sequence and the edited target image:
\begin{equation}
(\hat{T}_{\mathrm{tgt}}, \hat{I}_{\mathrm{tgt}})=f_{\theta}(I_{\mathrm{src}}, s, t).
\label{eq:formulation}
\end{equation}
Here, $\hat{T}_{\mathrm{tgt}}$ denotes the predicted target-language text sequence, and $\hat{I}_{\mathrm{tgt}}$ denotes the translated image with the source text rewritten in the target language. Although implemented as a single end-to-end model, UniTranslator couples translation understanding and visual text editing through shared multimodal representations, enabling semantic prediction and image generation to be optimized jointly within one unified framework.

\subsection{Overview of UniTranslator}

\begin{figure}[t]
\centering
\includegraphics[width=0.9\textwidth]{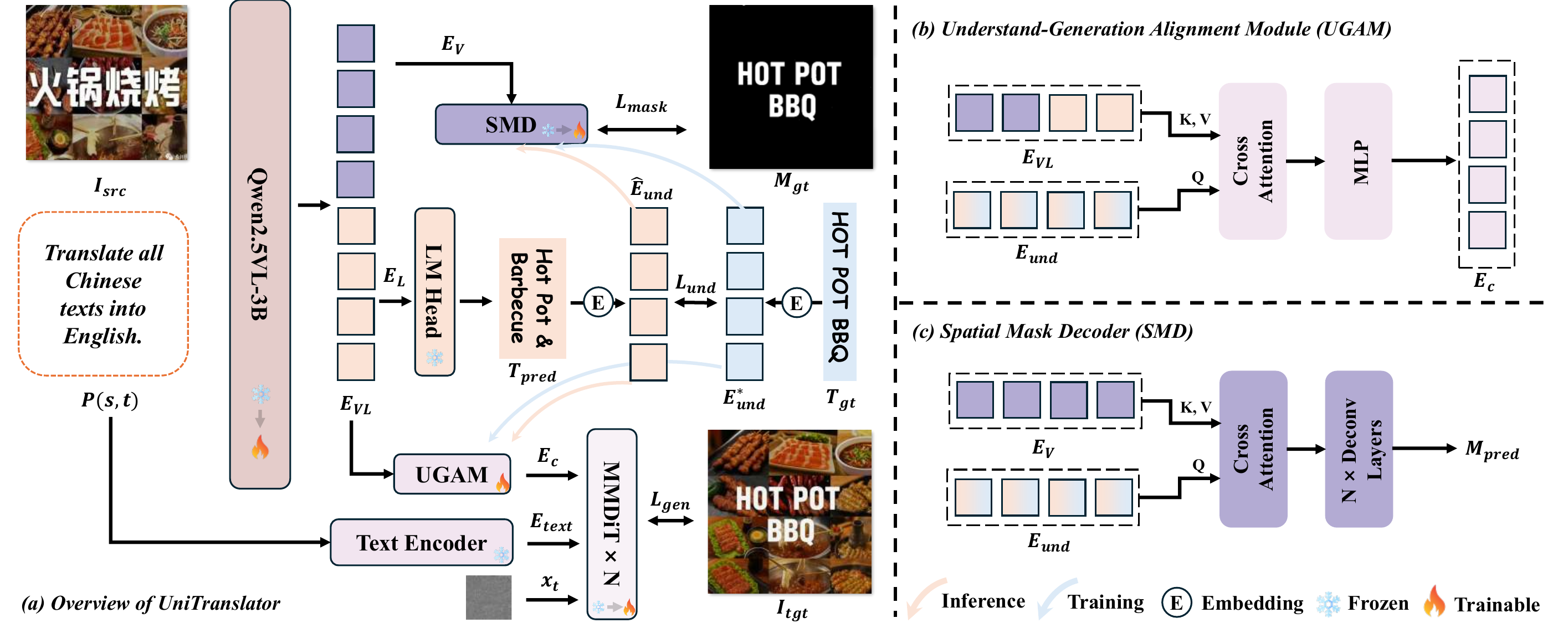}
\caption{Overview of UniTranslator, a unified multimodal framework for in-image machine translation. It introduces UGAM for understand-generation alignment and SMD for spatially grounded text editing.}
\label{fig:method}
\vspace{-15pt}
\end{figure}

As shown in \cref{fig:method}, UniTranslator is a unified multimodal model for in-image machine translation. Given a source image $I_{\mathrm{src}}$ and a translation instruction prompt $P(s,t)$, the model first encodes the visual-textual input with a multimodal backbone to obtain shared multimodal representations, which are then used by three task-specific branches. The \emph{understanding branch} predicts the translated text sequence $\hat{T}_{\mathrm{tgt}}$ through autoregressive decoding. The \emph{generation branch} synthesizes the translated image $\hat{I}_{\mathrm{tgt}}$ conditioned on the source image and aligned multimodal representations. The \emph{mask branch} predicts a text-region mask $\hat{M}$ to provide explicit spatial supervision for localized text rendering.

To address the two key challenges in unified multimodal IIMT, UniTranslator introduces two dedicated modules. First, an \textbf{Understand-Generation Alignment Module (UGAM)} bridges the understanding and generation branches by aligning the translation-related representations used for text prediction and image generation, thereby reducing semantic inconsistency between the two stages. Second, a \textbf{Spatial Mask Decoder (SMD)} provides pixel-level supervision over text regions, improving spatial grounding and layout-preserving text editing.
These components are jointly optimized within a unified framework. The overall training objective is defined as
\begin{equation}
\mathcal{L}_{\mathrm{total}}
=
\lambda_{\mathrm{und}} \mathcal{L}_{\mathrm{und}}
+
\lambda_{\mathrm{gen}} \mathcal{L}_{\mathrm{gen}}
+
\lambda_{\mathrm{mask}} \mathcal{L}_{\mathrm{mask}},
\label{eq:overall}
\end{equation}
where $\mathcal{L}_{\mathrm{und}}$, $\mathcal{L}_{\mathrm{gen}}$, and $\mathcal{L}_{\mathrm{mask}}$ denote the text prediction loss, the image generation loss, and the mask prediction loss, respectively.

\subsection{Understand-Generation Alignment Module}

To address the {Understanding and Generation Conflicts}, we introduce the \textbf{Understand Generation Alignment Module (UGAM)}. In unified multimodal IIMT, the understanding branch is optimized for autoregressive text prediction, whereas the generation branch requires a continuous conditioning representation for visual writeback, leading to a mismatch between the two objectives. As shown in \cref{fig:method}(b), UGAM takes the understanding representation ${E}{\mathrm{und}}$ as query and the multimodal representation $E{VL}$ as key and value, and applies cross-attention followed by an MLP to produce an aligned condition embedding $E_c$:
\begin{equation}
E_c = MLP(CrossAttn({E}{\mathrm{und}}, E{VL}, E_{VL})).
\label{eq:ugam}
\end{equation}
The resulting aligned embedding $E_c$ serves as a soft, generation-compatible condition. Instead of rigidly dictating what exact words to render, $E_c$ tells the generation branch (MMDiT) how the semantic translation aligns with the visual space, thereby mitigating the semantic-to-visual penalty clash. Concurrently, the understanding branch is trained with the standard autoregressive loss to maintain translation accuracy:
\begin{equation}
\mathcal{L}_{\mathrm{und}}
=
-\frac{1}{N}
\sum_{i=1}^{N}
\log p_{\theta}
\bigl(
T_{\mathrm{gt},i}\mid T_{\mathrm{gt},<i}, E_{VL}
\bigr),
\label{eq:l_und}
\end{equation}
where $T_{\mathrm{gt}}$ denotes the ground-truth translated text and $N$ is the number of target tokens. 
Ultimately, UGAM transforms $E_{\mathrm{und}}$ from a purely semantic representation into a robust, multi-modal bridge, ensuring semantic consistency without collapsing under strict pixel-level generative supervision.

\subsection{Spatial Mask Decoder}

Although UGAM improves semantic consistency between translation understanding and image generation, in-image machine translation further requires accurate spatial placement of translated text. 
To address this issue, we introduce a \textbf{Spatial Mask Decoder (SMD)} to predict the target text-region mask. Specifically, SMD takes the visual representation $E_V$ as key and value and the understanding representation ${E}{\mathrm{und}}$ as query. A cross-attention module first injects translation-aware semantics into spatial visual features, and the fused representation is then decoded to produce the target mask:
\begin{equation}
\hat{M}_{\mathrm{pred}} = 
Deconv(CrossAttn(({E}_{\mathrm{und}}, E_V, E_V))).
\label{eq:smd}
\end{equation}
We supervise the predicted mask $\hat{M}{\mathrm{pred}}$ with the ground-truth text-region mask $M_{\mathrm{gt}}$ using a combination of binary cross-entropy and Dice loss:
\begin{equation}
\mathcal{L}_{\mathrm{mask}}=
\lambda_{\mathrm{bce}} \mathcal{L}_{\mathrm{BCE}}(\hat{M}_{\mathrm{pred}}, M_{\mathrm{gt}})
+
\lambda_{\mathrm{dice}} \mathcal{L}_{\mathrm{Dice}}(\hat{M}_{\mathrm{pred}}, M_{\mathrm{gt}}).
\label{eq:l_mask}
\end{equation}
By introducing explicit pixel-level supervision on text regions, SMD improves spatial grounding and enables more accurate preservation of text position and layout during visual writeback.

\subsection{Conditional Image Generation}

Given the aligned condition produced by UGAM, UniTranslator employs a diffusion transformer to synthesize the translated image. Specifically, the generation branch takes the noised latent $x_t$ at diffusion step $t$, together with the aligned multimodal condition $E_c$ and the text condition $E_{\mathrm{text}}$, and predicts the corresponding denoising target:
\begin{equation}
L_{\mathrm{gen}}
=
\mathbb{E}_{(x_0, E_c, E_{\mathrm{text}})\sim\mathcal{D},\, t,\, \epsilon}
\left[
\left\|
v_{\theta}(x_t, t, E_c, E_{\mathrm{text}}) - v_t
\right\|_2^2
\right],
\label{eq:l_gen}
\end{equation}
where $x_0$ is the latent representation of the ground-truth translated image $I_{\mathrm{tgt}}$, and $v_t$ is the diffusion training target at step $t$.
In our framework, $E_c$ provides the aligned semantic condition transferred from the understanding branch through UGAM, while $E_{\mathrm{text}}$ is obtained from the instruction text encoder, as shown in \cref{fig:method}(a). Conditioning image generation on both signals allows the model to preserve the original visual content while rendering translated text that is semantically consistent with the predicted translation.

\subsection{Multi-Stage Training}




We adopt a two-stage training strategy to progressively adapt the unified multimodal model to in-image machine translation.

\noindent\textbf{Stage 1: Module Warm-up.}
In the first stage, we freeze the main pretrained backbones and optimize only the newly introduced task-specific modules, including UGAM and SMD. The multimodal backbone provides the visual and multimodal representations used by the understanding, alignment, and mask branches, while the diffusion generator remains fixed. This stage allows the model to establish a stable interface between translation understanding and image generation without disrupting the pretrained priors of the backbone and generator.

\noindent\textbf{Stage 2: Joint Fine-tuning.}
In the second stage, we then jointly fine-tune the understanding and generation pathways, including LoRA adapters in Qwen2.5-VL, UGAM, SMD, and the attention projection layers in MMDiT. This stage improves end-to-end instruction following, translation robustness, and image generation fidelity under realistic inference conditions. During training, UGAM uses the ground-truth translation $T_{\mathrm{gt}}$ as the text condition. During inference, UniTranslator instead uses the autoregressively predicted translation $T_{\mathrm{pred}}$. The second stage therefore also serves to reduce the distribution gap between teacher-forced supervision and autoregressive decoding, improving robustness at test time.

\section{Experiment}

\begin{table*}[t]
\centering
\small
\setlength{\tabcolsep}{4.0pt}
\caption{Results on German$\leftrightarrow$English image translation (De$\leftrightarrow$En) on the Translatotron-V benchmark. Best are in \textbf{bold} and \underline{underline}, respectively.}
\resizebox{\linewidth}{!}{%
\begin{tabular}{l c c c c c c}
\toprule
& \multicolumn{3}{c}{De$\rightarrow$En} & \multicolumn{3}{c}{En$\rightarrow$De} \\
\cmidrule(lr){2-4}\cmidrule(lr){5-7}
Model
& BLEU$\uparrow$ & Structure-BLEU$\uparrow$ & SSIM$\uparrow$
& BLEU$\uparrow$ & Structure-BLEU$\uparrow$ & SSIM$\uparrow$ \\
\midrule

\multicolumn{7}{c}{\textbf{Properiety Models}} \\ \midrule 
GPT-Image-1~\cite{gptimage1}        & \underline{20.75} & \underline{16.01} & 0.7369 & \textbf{19.61} & \textbf{17.44} & 0.7528 \\
Nano-Banana~\cite{comanici2025gemini}        & 10.16 & 10.41 & 0.7474 &  9.95 &  9.71 & 0.7488 \\
\midrule

\multicolumn{7}{c}{\textbf{Open-source Models}} \\ \midrule  
GLM-Image~\cite{glmimage}          &  0.40 &  0.40 & 0.6194 &  0.19 &  0.00 & 0.6038 \\
Qwen-Image~\cite{wu2025qwen}         &  0.23 &  0.23 & 0.7858 &  0.21 &  0.21 & 0.7862 \\
LongCat-Image~\cite{longcatimage}      &  0.22 &  0.22 & 0.7552 &  0.09 &  0.11 & 0.6925 \\
Janus~\cite{chen2025janus}              &   0.00  &   0.00  &   0.7202   &   0.00  &   0.00  &   0.7152   \\
Bagel~\cite{deng2025emerging}              &  0.18 &  0.18 & 0.7218 &  0.07 &  0.00 & 0.7063 \\
Uniworld~\cite{lin2025uniworld}           &  0.25 &  0.24 & \underline{0.7905} &  0.20 &  0.20 & \underline{0.7874} \\

\midrule

\multicolumn{7}{c}{\textbf{Specialist Model}} \\ \midrule 
OCR+MT+T2I                 & 15.37 & 14.87 & 0.7785 & 13.22 & 12.57 & 0.7550 \\
TIT+T2I                    & 14.80 & 14.73 & 0.7812 & 12.92 & 12.74 & 0.7620 \\
PEIT+T2I~\cite{zhu2023peit}                   & 10.91 & 10.78 & 0.7740 &  8.77 &  8.01 & 0.7594 \\
Pixel-level Transformer    &  0.15 &  0.15 & 0.7538 &  1.11 &  1.22 & 0.7616 \\
Translatotron-V~\cite{lan2024translatotron}            & 15.39 & 15.26 & 0.7832 & 13.23 & 12.92 & 0.7629 \\
UniTranslator (Ours)      & \textbf{25.03} & \textbf{24.86} & \textbf{0.8184} & \underline{13.41} & \underline{13.36} & \textbf{0.7887} \\
\bottomrule
\end{tabular}%
}
\label{tab:de_en_results}
\vspace{-20pt}
\end{table*}

\begin{table*}[t]
\centering
\small
\caption{Quantitative results on French/Romanian$\rightarrow$English image translation (Fr/Ro$\rightarrow$En) on the Translatotron-V benchmark. Best are in \textbf{bold} and \underline{underline}, respectively.}
\setlength{\tabcolsep}{4.0pt}
\resizebox{\linewidth}{!}{%
\begin{tabular}{l c c c c c c}
\toprule
& \multicolumn{3}{c}{Fr$\rightarrow$En} & \multicolumn{3}{c}{Ro$\rightarrow$En} \\
\cmidrule(lr){2-4}\cmidrule(lr){5-7}
Model
& BLEU$\uparrow$ & Structure-BLEU$\uparrow$ & SSIM$\uparrow$
& BLEU$\uparrow$ & Structure-BLEU$\uparrow$ & SSIM$\uparrow$ \\
\midrule

\multicolumn{7}{c}{\textbf{Properiety Models}} \\ \hline
GPT-Image-1~\cite{gptimage1}
& \underline{27.38} & \underline{23.05} & 0.7081 & \textbf{24.50} & \textbf{23.14} & 0.7386 \\
Nano-Banana~\cite{comanici2025gemini}        & 12.70 & 12.91 & 0.7495 &  9.30 &  8.73 & 0.7635 \\
\midrule

\multicolumn{7}{c}{\textbf{Open-source Models}} \\ \hline
GLM-Image ~\cite{glmimage}         &  0.20 &  0.00 & 0.6181 &  0.25 &  0.00 & 0.6278 \\
Qwen-Image ~\cite{wu2025qwen}        &  0.20 &  0.20 & 0.7767 &  0.22 &  0.22 & \underline{0.7908} \\
LongCat-Image ~\cite{longcatimage}     &  0.06 &  0.06 & 0.7451 &  0.06 &  0.07 & 0.7361 \\
Janus~\cite{chen2025janus}              &   0.00  &   0.00  &   0.7080   &   0.00  &   0.00  &   0.7194   \\
Bagel~\cite{deng2025emerging}              &  0.13 &  0.00 & 0.7239 &  0.10 &  0.00 & 0.7273 \\
Uniworld~\cite{lin2025uniworld}           &  0.13 &  0.13 & 0.7778 &  0.16 &  0.15 & 0.7906 \\
\midrule

\multicolumn{7}{c}{\textbf{Specialist Model}} \\ \hline 
OCR+MT+T2I                 & 21.60 & 21.58 & 0.7738 & 18.34 & 18.61 & 0.7752 \\
TIT+T2I                    & 21.87 & 21.78 & 0.7801 & 18.39 & 18.30 & 0.7764 \\
PEIT+T2I~\cite{zhu2023peit}                   & 18.51 & 18.55 & 0.7741 & 14.54 & 14.90 & 0.7704 \\
Pixel-level Transformer    &  2.08 &  2.61 & 0.7753 &  1.58 &  2.11 & 0.7696 \\
Translatotron-V ~\cite{lan2024translatotron}          & 22.20	& 22.17 & 	0.7811 &	18.44 &	\underline{18.73} &	0.7780 \\
UniTranslator      & \textbf{27.77} & \textbf{27.14} & \textbf{0.8060} & \underline{18.45} & 18.29 & \textbf{0.8045} \\
\bottomrule
\end{tabular}%
}
\label{tab:fr_ro_en_results}
\vspace{-15pt}
\end{table*}

\subsection{Settings}
We train UniTranslator in two stages using AdamW with a weight decay of $1\times10^{-4}$ and bf16 mixed precision. Gradient checkpointing and gradient accumulation with 8 steps are enabled throughout training. In the first stage, we freeze the pretrained Qwen2.5VL and diffusion backbone and optimize only the lightweight UGAM that connects the understanding and generation components. We use a constant learning rate of $1\times10^{-4}$ and train on $256\times256$ images with a per-device batch size of 32. In the second stage, we jointly fine-tune the trainable modules in Qwen2.5-VL with LoRA rank set to 64. We use a cosine learning-rate schedule with a peak learning rate of $1\times10^{-5}$ and 100 warm-up steps, and set the per-device batch size to 8. Stage 1 and Stage 2 are trained for 70 and 30 epochs, respectively. To balance the optimization objectives at a comparable scale, we set the loss weights to $\lambda_{\mathrm{und}}=0.1$, $\lambda_{\mathrm{gen}}=1$, and $\lambda_{\mathrm{mask}}=0.1$. Following common practice~\cite{wang2025marten}, we set $\lambda_{\mathrm{bce}}=1$ and $\lambda_{\mathrm{dice}}=1$. All experiments are conducted on NVIDIA H800 GPUs.
\subsection{Benchmarks}

\noindent\textbf{Translatotron-V(ision)}~\cite{lan2024translatotron} is a synthetic IIMT benchmark with relatively simple backgrounds and oriented text. It covers four languages (German, English, French, and Romanian) and is built from three public MT corpora: IWSLT14 German--English~\cite{cettolo2014report}, IWSLT17 French--English~\cite{cettolo2017overview}, and IWSLT14 Romanian--English~\cite{cettolo2017overview}. Following the official splits, the test sets contain 3,527, 4,927, and 1,008 image--translation pairs for German$\rightarrow$English, French$\rightarrow$English, and Romanian $\rightarrow$ English, respectively. All images are of size $512\times512$. We use EasyOCR\footnote{\url{https://github.com/JaidedAI/EasyOCR}} to extract translated text from generated images, and report BLEU~\cite{papineni2002bleu}, structure-BLEU~\cite{lan2024translatotron}, and SSIM~\cite{wang2004image} for evaluation.

\noindent\textbf{IIMT30k}~\cite{tian2025exploring} is a more realistic benchmark for instance-level IIMT, where synthetic text is rendered onto real-world photographs. Built on Multi30k~\cite{DBLP:conf/acl/ElliottFSS16}, it uses real images as backgrounds and image captions as textual content, resulting in cluttered scenes with diverse textures and illumination. The dataset includes German and English. Following the official splits, the validation and test sets contain 864 and 2,740 samples, respectively. All images are resized and cropped to $48\times512$ while preserving the full text region and surrounding context. We evaluate translation quality using EasyOCR followed by BLEU and COMET~\cite{rei-etal-2020-comet}, and report FID\footnote{\url{https://github.com/mseitzer/pytorch-fid}} for image quality.

\noindent\textbf{PRIM}~\cite{tian2025prim} is a multilingual IIMT benchmark designed for practical scenarios with diverse backgrounds, fonts, placements, and translation directions. It provides a large-scale synthetic training set, MTedIIMT, constructed by pairing multilingual parallel text with real-world TED imagery and rendering text using TRDG\footnote{\url{https://github.com/Belval/TextRecognitionDataGenerator}}. PRIM contains 7M training samples and 3K test samples. We follow the official protocol and report BLEU, COMET, and FID for evaluation.

\subsection{Results and Discussion}

\begin{table*}[t]
\centering
\small
\setlength{\tabcolsep}{5.5pt}
\caption{Experimental results of different systems. Metrics include translation quality (BLEU, COMET(\%)) and visual effect (FID). $\uparrow$/$\downarrow$ indicates higher/lower is better. Best and second-best are in \textbf{bold} and \underline{underline}, respectively.}
\resizebox{\linewidth}{!}{%
\begin{tabular}{l c c c c c c c c}
\toprule
\multirow{3}{*}{\textbf{Systems}} &
\multicolumn{4}{c}{\textbf{Translation Quality (BLEU}$\uparrow$ / \textbf{COMET}$\uparrow$\textbf{)}} &
\multicolumn{4}{c}{\textbf{Visual Effect (FID}$\downarrow$\textbf{)}} \\
\cmidrule(lr){2-5}\cmidrule(lr){6-9}
& \multicolumn{2}{c}{De-En} & \multicolumn{2}{c}{En-De}
& \multicolumn{2}{c}{De-En} & \multicolumn{2}{c}{En-De} \\
\cmidrule(lr){2-3}\cmidrule(lr){4-5}\cmidrule(lr){6-7}\cmidrule(lr){8-9}
& Valid & Test & Valid & Test & Valid & Test & Valid & Test \\
\midrule
VQGAN~\cite{Esser_2021_CVPR}
& 0.7 / 24.2 & 0.6 / 24.4 & 0.6 / 17.9 & 0.8 / 20.2
& 25.5 & 21.3 & 27.2 & 20.7 \\
VAR~\cite{tian2024visual}
& 0.1 / 22.5 & 0.1 / 22.2 & 0.3 / 18.3 & 0.2 / 17.7
& {21.2} & 15.7 & {20.1} & {14.1} \\
MaskGIT~\cite{chang2022maskgit}
& 0.2 / 23.7 & 0.2 / 24.0 & 0.1 / 18.9 & 0.2 / 18.5
& 21.5 & 20.9 & 22.2 & 18.8 \\
TIT-Render~\cite{vaswani2023attention}
& 13.8 / {53.1} & {12.1} / {49.6} & \underline{14.0} / \textbf{46.3} & 10.2 / \underline{43.9}
& 137.4 & 133.2 & 121.7 & 119.0 \\
PEIT-Render~\cite{zhu2023peit}
& 10.5 / 50.4 & 8.6 / 47.0 & 12.3 / 41.1 & 7.9 / 35.6
& 149.9 & 140.1 & 125.3 & 119.7 \\
McTIT-Render~\cite{lan2023exploring}
& {14.2} / \underline{53.5} & 11.7 / 47.3 & 13.5 / {43.3} & {10.5} / {42.8}
& 141.2 & 137.5 & 124.9 & 117.5 \\
Translatotron-V~\cite{lan2024translatotron}
& 2.7 / 30.1 & 1.6 / 24.8 & 2.1 / 26.5 & 1.9 / 24.6
& 22.4 & {10.1} & 23.2 & 17.5 \\
DebackX~\cite{tian2025exploring}
& \underline{14.9} / 51.2 & \underline{12.8} / \underline{50.0} & \textbf{14.6} / 42.2 & \underline{11.1} / 40.0
& \underline{20.4} & \underline{9.0} & \underline{19.5} & \textbf{8.7} \\
UniTranslator & \textbf{16.3} / \textbf{59.9} & \textbf{14.7} / \textbf{59.8} & 13.1 / \underline{45.0} & \textbf{13.0} / \textbf{45.5} & \textbf{18.6} & \textbf{8.9} & \textbf{19.2} & \underline{12.5} \\
\bottomrule
\end{tabular}%
}
\label{tab:iimt30k}
\vspace{-10pt}
\end{table*}

\begin{table*}[t]
\centering
\small
\setlength{\tabcolsep}{6pt}
\caption{Experimental results on PRIM. Metrics include translation quality (BLEU, COMET(\%)) and visual effect (FID), and ``Avg.'' represents the average across all translation directions. $\uparrow$ or $\downarrow$ indicates that higher or lower values are better. The best and second-best performances are in bold and underline, respectively.}
\resizebox{\linewidth}{!}{%
\begin{tabular}{l c c c c c c | c}
\toprule
\multirow{2}{*}{\textbf{Systems}} &
\multicolumn{6}{c}{\textbf{BLEU}$\uparrow$ / \textbf{COMET}$\uparrow$} &
\multicolumn{1}{c}{\textbf{FID}$\downarrow$} \\
\cmidrule(lr){2-7}\cmidrule(lr){8-8}
& En-De & En-Fr & En-Cs & En-Ru & En-Ro & Avg. & Avg. \\
\midrule
\multicolumn{8}{c}{\textbf{Cascade Models}} \\
\midrule
PARSeq-mTransformer-Render~\cite{bautista2022scene}
& 9.5 / 41.7
& 13.8 / 46.9 
& \underline{7.7} / \underline{43.9}
& {5.5} / {48.1}
& {12.8} / \textbf{53.5}
& 9.9 / 46.8
& 103.8 \\
PEIT-Render~\cite{zhu2023peit}
& {10.4} / \underline{45.1}
& {14.0} / {48.1}
& \textbf{7.9} / \textbf{46.2}
& 5.3 / 47.7
& \textbf{14.2} / \underline{52.9}
& {10.4} / \underline{48.0}
& 101.4 \\
\midrule
\multicolumn{8}{c}{\textbf{End-to-end Models}} \\
\midrule
TranslatotronV~\cite{lan2024translatotron}
& 1.7 / 34.3 & 1.9 / 30.2 & 1.1 / 30.5 & 0.9 / 32.0 & 1.3 / 33.9 & 1.4 / 32.2 & 69.1 \\
VisTrans~\cite{tian2025prim} & \underline{12.6} / {44.4}
& \underline{17.0} / {49.4}
& 5.9 / 41.8
& \underline{7.2} / \underline{49.4}
& \underline{13.9} / 50.2
& \underline{11.3} / {47.0}
& \underline{28.8} \\
UniTranslator &  \textbf{13.1} / \textbf{48.7}
& \textbf{21.6} / \textbf{58.5}
& {7.4} / 43.3
& \textbf{8.4} / \textbf{50.2}
& 13.5 / {52.8}
& \textbf{12.8} / \textbf{50.7}
& \textbf{22.9} \\
\bottomrule
\end{tabular}%
}
\label{tab:prim}
\end{table*}

\noindent\textbf{Translatotron-V(ision).}
We first evaluate UniTranslator on Translatotron-V(ision), which contains four subtasks: De$\rightarrow$En, En$\rightarrow$De, Fr$\rightarrow$En, and Ro$\rightarrow$En. As shown in \cref{tab:de_en_results,tab:fr_ro_en_results}, existing open-source unified multimodal models perform poorly in the zero-shot setting, while strong closed-source models such as GPT-Image-1 already show competitive translation performance, indicating the potential of unified multimodal models for IIMT. However, these models still lag behind specialist systems in visual faithfulness and editing precision, as reflected by lower SSIM and a larger gap between BLEU and structure-BLEU. In contrast, UniTranslator achieves the strongest overall results across all four subtasks, substantially improving BLEU and structure-BLEU while maintaining competitive or better SSIM, showing that it preserves both translation accuracy and spatially faithful text rendering.

\noindent\textbf{IIMT30k.}
On the more realistic IIMT30k benchmark, UniTranslator achieves the best overall performance on both translation quality and visual quality (\cref{tab:iimt30k}). It obtains the highest BLEU and COMET scores in both De$\rightarrow$En and En$\rightarrow$De, and also achieves the lowest FID by a clear margin. The especially large gain in FID suggests that UniTranslator is better suited to realistic scene editing, where cluttered backgrounds and complex textures make faithful visual generation more critical.

\noindent\textbf{PRIM.}
On PRIM, a large-scale multilingual benchmark with more diverse scenes and translation directions, UniTranslator achieves the best average performance among end-to-end models, with the strongest overall BLEU/COMET and the lowest FID (\cref{tab:prim}). Although it does not always achieve the best translation score on every individual direction, likely due to the limited support of current foundation models for some lower-resource languages, it provides the best overall balance between translation quality and visual realism. The particularly large FID gain further shows the advantage of UniTranslator in practical multilingual scenarios that are closer to real-world image editing.

\subsection{Ablations}

\begin{table}[t]
\centering
\caption{Ablation of UGAM and SMD in UniTranslator on Translatotron-V (De$\rightarrow$En). Structured-BLEU and SSIM are reported, where higher is better.}
\label{tab:ablation_modules}
\begin{tabular}{lcccc}
\hline
Method & UGAM & SMD  & Structured-BLEU $\uparrow$ & SSIM $\uparrow$ \\
\hline
Baseline                 & $\times$ & $\times$ & 9.00 & 0.8081 \\
\# 1 & $\checkmark$ & $\times$ & 14.28  &  0.8036 \\
\# 2         & $\times$ & $\checkmark$ & 16.02 & 0.8138\\
UniTranslator         & $\checkmark$ & $\checkmark$ & \textbf{24.86} & \textbf{0.8184} \\
\hline
\end{tabular}
\vspace{-10pt}
\end{table}

\begin{table}[t]
\centering
\caption{Ablation of the two-stage training strategy on Translatotron-V (De$\rightarrow$En). $\Delta$ denotes the change relative to the full model. Higher BLEU and SSIM are better.}
\label{tab:ablation_multistage_delta}
\begin{tabular}{lcccc}
\hline
Settings & Stage 1 & Stage 2 & BLEU $\uparrow$ & SSIM $\uparrow$ \\
\hline
Ours      & $\checkmark$ & $\checkmark$     & \textbf{25.03} & \textbf{0.8184} \\
\# 1     & $\checkmark$ & & 12.53 ($\Delta$ -12.50)  & 0.7782 ($\Delta$ -0.0402)  \\
\# 2    & & $\checkmark$ & 15.08 ($\Delta$ -9.95) & 0.8148 ($\Delta$ -0.0036) \\
\hline
\end{tabular}
\label{tab:stage}
\vspace{-15pt}
\end{table}

We conduct ablation studies to examine the contributions of the two main components in UniTranslator, namely UGAM and SMD, as well as the proposed two-stage training strategy.

\noindent\textbf{Module Ablations.}
As shown in \cref{tab:ablation_modules}, removing both UGAM and SMD leads to a clear performance drop, indicating that naïve joint training is insufficient for reliable in-image translation. Enabling UGAM alone mainly improves translation quality, showing that consistent writeback supervision is critical for stabilizing the understanding branch and providing effective text conditions for generation. In contrast, enabling SMD alone improves both structure-aware translation quality and image fidelity, confirming that explicit spatial supervision helps localize text regions and suppress imprecise edits. Combining the two yields the best performance by a large margin, demonstrating that UGAM and SMD play complementary roles: UGAM improves semantic consistency, while SMD strengthens spatial accuracy.

\noindent\textbf{Stage Ablations.}
\cref{tab:ablation_multistage_delta} validates the necessity of the two-stage training paradigm. Using only Stage~1 results in poor performance, suggesting that alignment warm-up alone is insufficient for robust end-to-end generation. Using only Stage~2 performs better, especially in SSIM, but still falls far behind the full model in BLEU, indicating that direct joint fine-tuning cannot fully establish a stable understanding-to-generation interface. The full two-stage training achieves the best overall results, showing that Stage~1 is important for learning a reliable cross-modal conditioning bridge, while Stage~2 is necessary for end-to-end adaptation under realistic inference conditions.

\subsection{Synergy Between Generation and Understanding}

\cref{tab:synergy} studies the bidirectional interaction between understanding and generation in UniTranslator. Unlike conventional multi-task learning, where one branch mainly acts as auxiliary supervision, our goal is to examine whether the two capabilities provide mutual benefits under joint optimization. For \emph{understanding improves generation} (U$\rightarrow$G), adding the understanding loss $\mathcal{L}{\mathrm{und}}$ to generation training consistently improves Structure-BLEU and SSIM, indicating that language-side supervision provides stronger semantic conditions for text rendering and visual consistency. For the reverse direction \emph{generation improves understanding} (G$\rightarrow$U), jointly optimizing with the generation loss $\mathcal{L}{\mathrm{gen}}$ improves both BLEU and COMET, suggesting that the generation objective also enhances translation understanding by encouraging visually grounded and context-faithful representations. Overall, \cref{tab:synergy} shows that understanding and generation in UniTranslator form a bidirectionally reinforcing loop, supporting our key design choice that IIMT should be modeled as a unified understanding-and-generation problem rather than a loosely coupled pipeline.

\begin{table}[t]
\centering
\caption{Empirical experiments on the mutual synergy between generation and understanding. For the generation task, metrics A and B correspond to Structure-BLEU and SSIM, respectively; for the understanding task, metrics A and B report BLEU and COMET (\%). All experiments are conducted on the Translatotron-V dataset.}
\label{tab:abstract_metrics}
\begin{tabular}{@{}llccc@{}}
\toprule
 & & & \multicolumn{2}{c}{\textbf{Performance Metrics}} \\
\cmidrule(l){4-5}
\textbf{Hypothesis} & \textbf{Setup} & \textbf{Training Loss} & \textbf{Metric A ↑} & \textbf{Metric B ↑} \\
\midrule

\multirow{2}{*}{\begin{tabular}[c]{@{}l@{}}U improves G\\ \footnotesize{(Eval on Generation)}\end{tabular}}
    & Baseline   & $L_{\text{gen}}$                 & 16.57 & 0.8142 \\
    & Full Model & $L_{\text{gen}} + L_{\text{und}}$ & \textbf{25.03} & \textbf{0.8184} \\
\addlinespace
\midrule
\addlinespace
\multirow{2}{*}{\begin{tabular}[c]{@{}l@{}}G improves U\\ \footnotesize{(Eval on Understanding)}\end{tabular}}
    & Baseline   & $L_{\text{und}}$                 & 25.70 & 80.84 \\
    & Full Model & $L_{\text{und}} + L_{\text{gen}}$ & \textbf{34.71} & \textbf{84.61} \\

\bottomrule
\end{tabular}
\label{tab:synergy}
\end{table}

\begin{figure}[t]
\centering
\includegraphics[width=\textwidth]{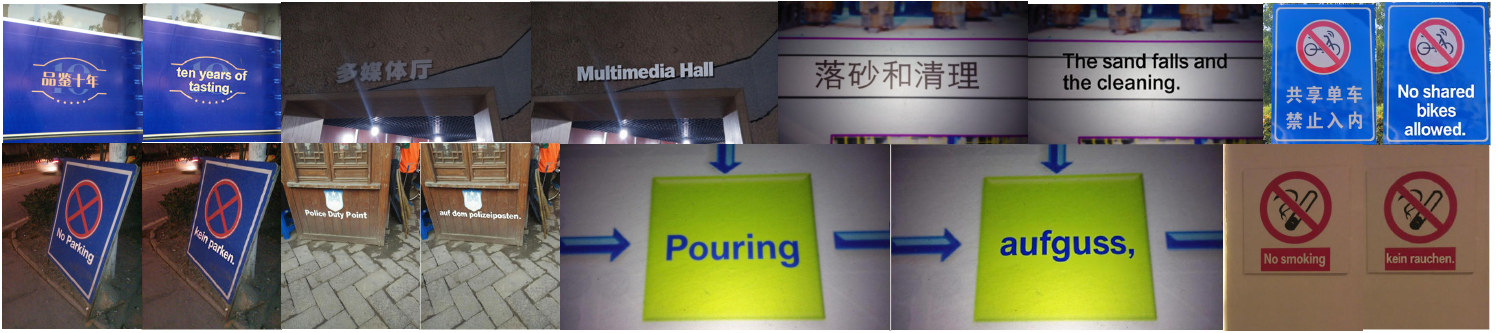}
    \caption{
    Qualitative results of zero-shot inference on the AnyTrans \cite{qian2024anytrans} dataset. The first row presents Chinese-to-English translation, while the second row presents English-to-German translation.
}
    \label{fig:vis_zs}
\end{figure}

\begin{figure}[t]
\centering
\includegraphics[width=0.9\textwidth]{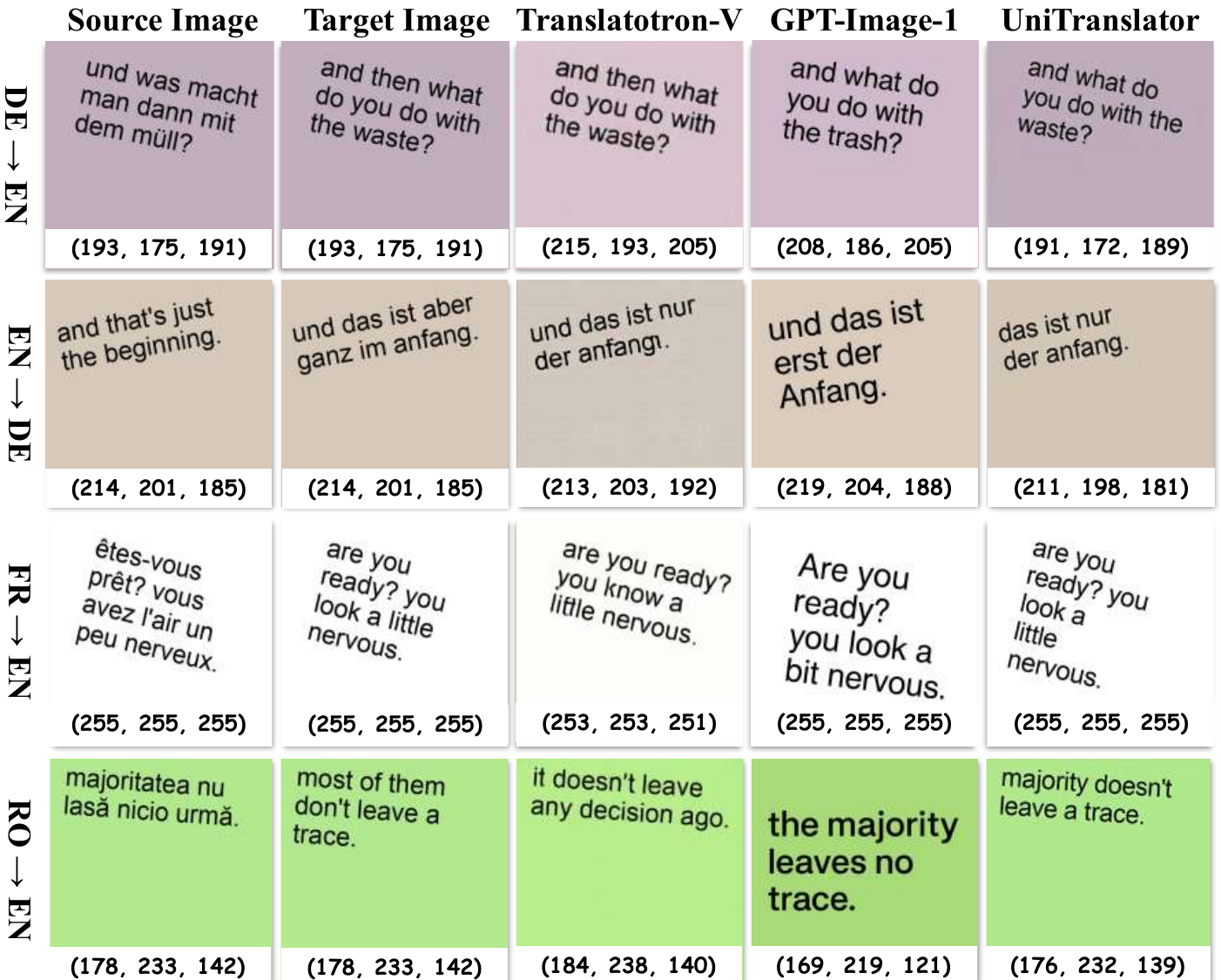}
    \caption{
Qualitative comparison on the Translatotron-V benchmark. The RGB values shown below each image correspond to the background color, providing a direct comparison of the predicted background colors produced by different models.
}
    \label{fig:vis}
    \vspace{-10pt}
\end{figure}

\section{Visualization \& Discussion}

~\cref{fig:vis} presents qualitative comparisons on the Translatotron-V benchmark across four subsets from top to bottom. From left to right, we show the source image, target image, the previous state-of-the-art Translatotron-V, GPT-Image-1, and our UniTranslator. Overall, UniTranslator achieves a strong balance between translation correctness, layout preservation, and source-faithful visual rendering. Compared with Translatotron-V, it more consistently preserves the background color and overall visual appearance of the source image, while avoiding noticeable background shifts in many cases. Compared with GPT-Image-1, it better maintains spatial consistency, with translated text more accurately aligned to the original position and layout.

We further observe that UniTranslator remains robust on images with cluttered backgrounds and diverse text layouts, and this advantage also transfers to zero-shot evaluation on the AnyTrans dataset in ~\cref{fig:vis_zs}. At the same time, we note that highly challenging cases are still not fully solved. In particular, under highly stylized typography, such as eroded graffiti or neon glowing cursive text, the model may fail to faithfully preserve fine-grained font attributes, and under extremely complex backgrounds it may still introduce unintended visual changes. Additional qualitative comparisons, limitation analysis, and failure cases are provided in the appendix.

\section{Conclusion}
We presented UniTranslator, a unified multi-modal framework for in-image machine translation that jointly optimizes translation understanding and image generation. To resolve the semantic conflict between language-level equivalence and pixel-level supervision, we introduced an Understand-Generation Alignment Module, a two-stage training pipeline for progressive alignment, and a mask-supervised refinement head for spatially precise text editing. Extensive experiments show that these components are all effective and complementary, and that the proposed training strategy is critical for robust performance. UniTranslator achieves state-of-the-art results on multiple benchmarks with improved cross-language and cross-scene generalization, while our analysis further reveals a consistent mutual benefit between understanding and generation. These results suggest that unified optimization is a promising direction for visually grounded text translation and editing.

\section*{Acknowledgement}

Supported by the National Natural Science Foundation of China (Grant NO 62376266 and 62406318). 
\clearpage  

%
%
\bibliographystyle{splncs04}
\bibliography{main}

\clearpage
\input{appendix}

\end{document}

%% file: appendix.tex
\appendix

The appendix includes the following aspects:
\begin{itemize}
    \item \ref{sec:efficiency_generalization}: Inference Efficiency and Task Generalization.
    \item \ref{sec:appendix_failure_cases}: Additional Failure Case Analysis. 
    \item \ref{sec:sota}: Details of Compared methods.
    \item \ref{sec:sbleu}: Details of the Structure-BLEU.
    \item \ref{sec:vis}:  More Visualization
\end{itemize}

\section{Inference Efficiency and Task Generalization}
\label{sec:efficiency_generalization}

We further evaluate UniTranslator in terms of inference efficiency and generalization to related localized editing tasks.

\paragraph{Inference efficiency.}
Table~\ref{tab:efficiency} compares the test-time latency and peak GPU memory usage of different methods under the same hardware setting, image resolution, and batch size. The cascaded OCR+MT+T2I pipeline is the most efficient, while Translatotron-V has the highest latency. UniTranslator reduces latency substantially compared with Translatotron-V (9.51 vs.\ 17.25 sec/img), showing that our unified framework remains practically efficient. At the same time, UniTranslator requires substantially higher GPU memory (50G), mainly due to the large multimodal backbone and generation module, which we leave for future optimization.

\paragraph{Task generalization.}
To verify that UGAM and SMD are not limited to IIMT, we conduct a proof-of-concept experiment on TextEditBench, a scene text editing benchmark that also requires instruction understanding, localized modification, and background preservation. As shown in Table~\ref{tab:generalization}, both UGAM and SMD consistently improve the baseline across all image fidelity metrics, and their combination achieves the best overall performance. This suggests that the proposed design addresses general challenges in unified understanding-and-generation tasks beyond IIMT.

\begin{table}[ht]
\centering
\small
\caption{Inference efficiency comparison.}
\begin{tabular}{lcc}
\toprule
Method & GPU Mem. & Sec./Img. $\downarrow$ \\
\midrule
OCR+MT+T2I (Cascaded) & 1.8G & 3.22 \\
Translatotron-V (E2E) & 3.4G & 17.25 \\
UniTranslator & 50G & 9.51 \\
\bottomrule
\end{tabular}
\label{tab:efficiency}
\end{table}

\section{Additional Failure Case Analysis}
\label{sec:appendix_failure_cases}

We provide additional qualitative analysis of the failure cases discussed in the rebuttal. Although UniTranslator performs well in translation correctness, layout preservation, and source-faithful rendering, challenging cases remain under highly stylized typography and complex backgrounds.

\begin{figure}[t]
    \centering
    \includegraphics[width=0.95\linewidth]{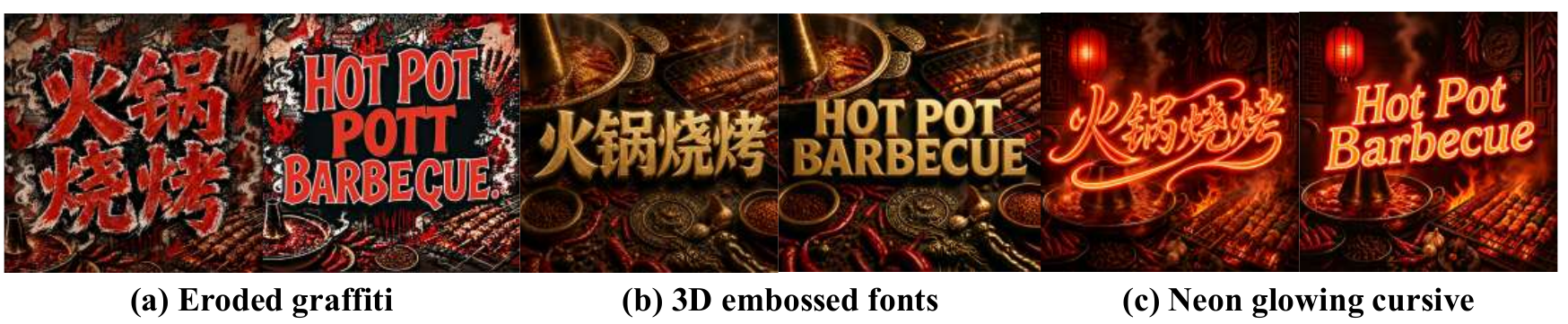}
    \caption{Failure cases on highly stylized typography. UniTranslator can usually localize the target region and generate semantically correct translated content, but may fail to fully preserve fine-grained font attributes such as stroke erosion, glow intensity, or cursive deformation.}
    \label{fig:appendix_failure_typography}
\end{figure}

\begin{figure}[t]
    \centering
    \includegraphics[width=0.95\linewidth]{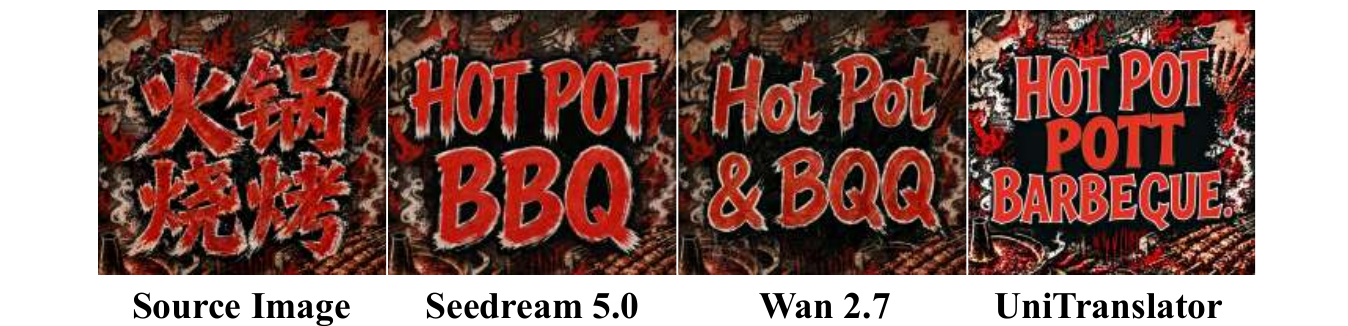}
    \caption{Failure cases on images with complicated backgrounds. While UniTranslator generally preserves the overall scene structure and text layout, it may still introduce unintended changes in background texture, color, or local appearance in difficult cases.}
    \label{fig:appendix_failure_background}
\end{figure}

As shown in Fig.~\ref{fig:appendix_failure_typography}, UniTranslator can usually localize the target text region and generate semantically correct translated content even for difficult styles such as eroded graffiti and neon glowing cursive fonts. However, it may fail to fully preserve fine-grained font attributes, such as stroke erosion, glow intensity, or cursive deformation. Fig.  ~\ref{fig:appendix_failure_background} further shows that, although the model generally preserves the overall scene structure, it may still introduce unintended edits outside the target text region in cluttered real-world scenes, including background texture changes, color deviations, or local appearance inconsistency. These observations highlight the need for better fine-grained style preservation and stronger background fidelity in future work.

\begin{table}[t]
\centering
\small
\caption{Generalization results on TextEditBench.}
\begin{tabular}{lcccc}
\toprule
Method & SSIM $\uparrow$ & LPIPS $\downarrow$ & PSNR $\uparrow$ & MSE $\downarrow$ \\
\midrule
Baseline & 0.8709 & 0.0864 & 20.69 & 1034 \\
+ UGAM & 0.8928 & 0.0695 & 23.08 & 816 \\
+ SMD & 0.8920 & 0.0697 & 23.11 & 815 \\
UniTranslator & 0.8937 & 0.0683 & 23.17 & 796 \\
\bottomrule
\end{tabular}
\label{tab:generalization}
\end{table}


\section{Details of Compared methods.}
\label{sec:sota}

\noindent\textbf{OCR+MT+T2I} follows a three-stage pipeline consisting of an OCR model, a machine translation (MT) model, and a text-to-image (T2I) model. For the OCR stage, we adopt a ViT-B encoder and a Transformer decoder following prior work. The decoder uses the Transformer-Base configuration. For the MT stage, we use a standard Transformer-Base model. The T2I stage includes a Transformer-based text encoder, a ResNet-based image encoder, an image decoder, and an image tokenizer. Both the text encoder and the image decoder use Transformer-Base settings, while the image encoder adopts a ResNet-50 backbone. The image tokenizer is configured identically to that used in our method. For fairness, the teacher model in our framework uses the same T2I architecture, except that the hidden size is reduced from 512 to 384.

\noindent\textbf{TIT+T2I} is constructed by cascading a text-in-image translation (TIT) model with the same T2I model described above. The TIT model consists of an image encoder, a source-text decoder, and a target-text decoder. The image encoder uses a ViT-B architecture, and both text decoders follow the Transformer-Base configuration.

\noindent\textbf{PEIT+T2I}~\cite{zhu2023peit} replaces the TIT model in TIT+T2I with PEIT. To ensure a fair comparison, we reimplement PEIT within our training and evaluation framework, scale the model to match our parameter budget, and do not use additional external data in its multi-stage training.

\noindent\textbf{Pixel-level Transformer} This model the same structure as Translatotron-V~\cite{lan2024translatotron} but removes the image tokenizer and directly predicts pixel values. It is trained using multi-task learning as well, with the IIMT task being optimized with a mean squared error loss due to the pixel values being of floating-point type.

\begin{algorithm}[ht]
\caption{Structure-BLEU}
\label{algorithm:structure-bleu}
\begin{algorithmic}[1]
\REQUIRE Generated image $\hat{\mathbf{t}}$, reference image $\mathbf{t}$

\STATE $\mathcal{R} \leftarrow \text{OCR}(\mathbf{t})$ \COMMENT{Set of reference texts and bounding boxes}
\STATE $\mathcal{H} \leftarrow \text{OCR}(\hat{\mathbf{t}})$ \COMMENT{Set of generated texts and bounding boxes}
\STATE $\mathcal{M} \leftarrow \{\}$ \COMMENT{Set of matched text pairs}

\FOR{each $h \in \mathcal{H}$}
    \STATE $m \leftarrow \text{None}$
    \STATE $s \leftarrow 0$
    \FOR{each $r \in \mathcal{R}$}
        \STATE $\hat{s} \leftarrow \text{IOU}(h, r)$
        \IF{$\hat{s} > s$}
            \STATE $s \leftarrow \hat{s}$
            \STATE $m \leftarrow (h, r)$
        \ENDIF
    \ENDFOR
    
    \IF{$s \geq 0.5$}
        \STATE $\mathcal{M} \leftarrow \mathcal{M} \cup \{m\}$
    \ENDIF
\ENDFOR
\RETURN BLEU($\mathcal{M}$)
\end{algorithmic}
\end{algorithm}

\noindent\textbf{Translatotron-V} \cite{lan2024translatotronv}. 
An end-to-end in-image machine translation model that directly generates the target image from the source image. Specifically, it introduces a target text decoder to alleviate the language alignment burden and an image tokenizer to convert pixels into shorter discrete visual tokens, enabling the model to generate RGB images in an end-to-end manner.

\noindent\textbf{VQGAN} \cite{esser2021taming}. 
A codebook-based image generation system that tokenizes images into discrete visual tokens and generates target images with an autoregressive Transformer decoder. In our experiments, we train the visual tokenizer on all source and target images, and then train the autoregressive decoder for each translation direction separately. 

\noindent\textbf{VAR} \cite{tian2024visual}.
An autoregressive image generation model built upon vector-quantized visual representations, which improves generation efficiency by adopting a next-scale prediction paradigm. We use it as an image-to-image generation baseline for translating source images directly into target images. 

\noindent\textbf{MaskGIT} \cite{chang2022maskgit}.
A mask-based image generation system that employs a bidirectional Transformer to predict masked visual tokens and iteratively refine the generated image at inference time. In our experiments, we adapt it as an image-to-image generation baseline for the in-image translation task. 

\noindent\textbf{TIT-Render} denotes the TIT+T2I pipeline.

\noindent\textbf{PEIT-Render} \cite{zhu2023peit}. 
A pipeline system that combines the pre-trained end-to-end text-image translation model PEIT with a text rendering module. The PEIT model is first used to generate the target-language text, and then the translated text is rendered into the image to obtain the final output. 

\noindent\textbf{MeTIT-Render} \cite{lan2023exploring} 
A pipeline system that combines the end-to-end text-image translation model MeTIT with a text rendering module. The MeTIT model predicts the translated text from the source image, after which a rendering module inserts the translated text back into the image. 

\noindent\textbf{DebackX} \cite{tian2025exploring}.
An end-to-end in-image machine translation model that explicitly decomposes the source image into background and text-image regions, translates the text-image component, and then fuses the translated text-image with the background to form the final target image. We use the original implementation as a strong end-to-end baseline. 

\noindent\textbf{ParSeq-mTransformer-Render} \cite{bautista2022scene}. 
A cascade baseline consisting of the state-of-the-art OCR model PARSeq, a multilingual machine translation model mTransformer, and a text rendering module. The OCR model first recognizes the source text in the image, the MT model translates the recognized text, and the rendering module writes the translated text back into the image. 

\noindent\textbf{VisTrans} \cite{tian2025prim}.
An end-to-end in-image machine translation model designed for practical scenarios, which processes visual text and background information separately to improve both translation quality and visual fidelity. We adopt it as a recent strong end-to-end baseline for comparison.

\section{Details of the Structure-BLEU.}
\label{sec:sbleu}
Structure-BLEU (S-BLEU) is proposed by Translatotron-V~\cite{lan2024translatotron}, aiming to evaluate text location information
for IIMT. S-BLEU extends the conventional BLEU metric by incorporating spatial structure information. Instead of directly comparing text sequences, S-BLEU first extracts text instances and their bounding boxes from both generated and reference images using OCR. Each generated text is matched with the reference text that has the highest IoU, and pairs with IoU above a predefined threshold are retained. The BLEU score is then computed over the matched text pairs. This design enables S-BLEU to evaluate not only textual similarity but also structural alignment between generated and reference images. ~\cref{algorithm:structure-bleu} is the detailed implementation.

\section{More Visualization.}
\label{sec:vis}

\begin{figure}[t]
\centering
\includegraphics[width=\textwidth]{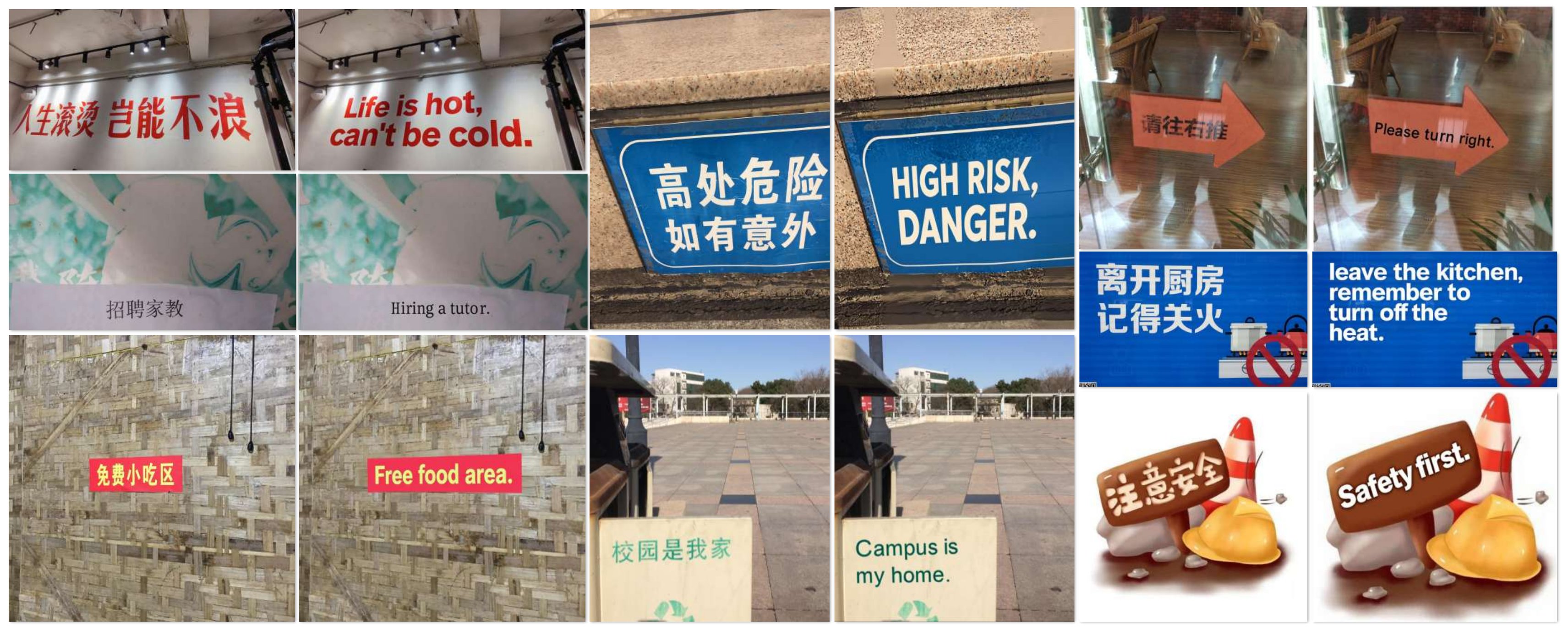}
    \caption{
Qualitative comparison on the AnyTrans benchmark~\cite{qian2024anytrans} in zero-shot settings. 
}
    \label{fig:vis_zs_anytrans}
\end{figure}

\begin{figure}[ht]
\centering
\includegraphics[width=0.8\textwidth]{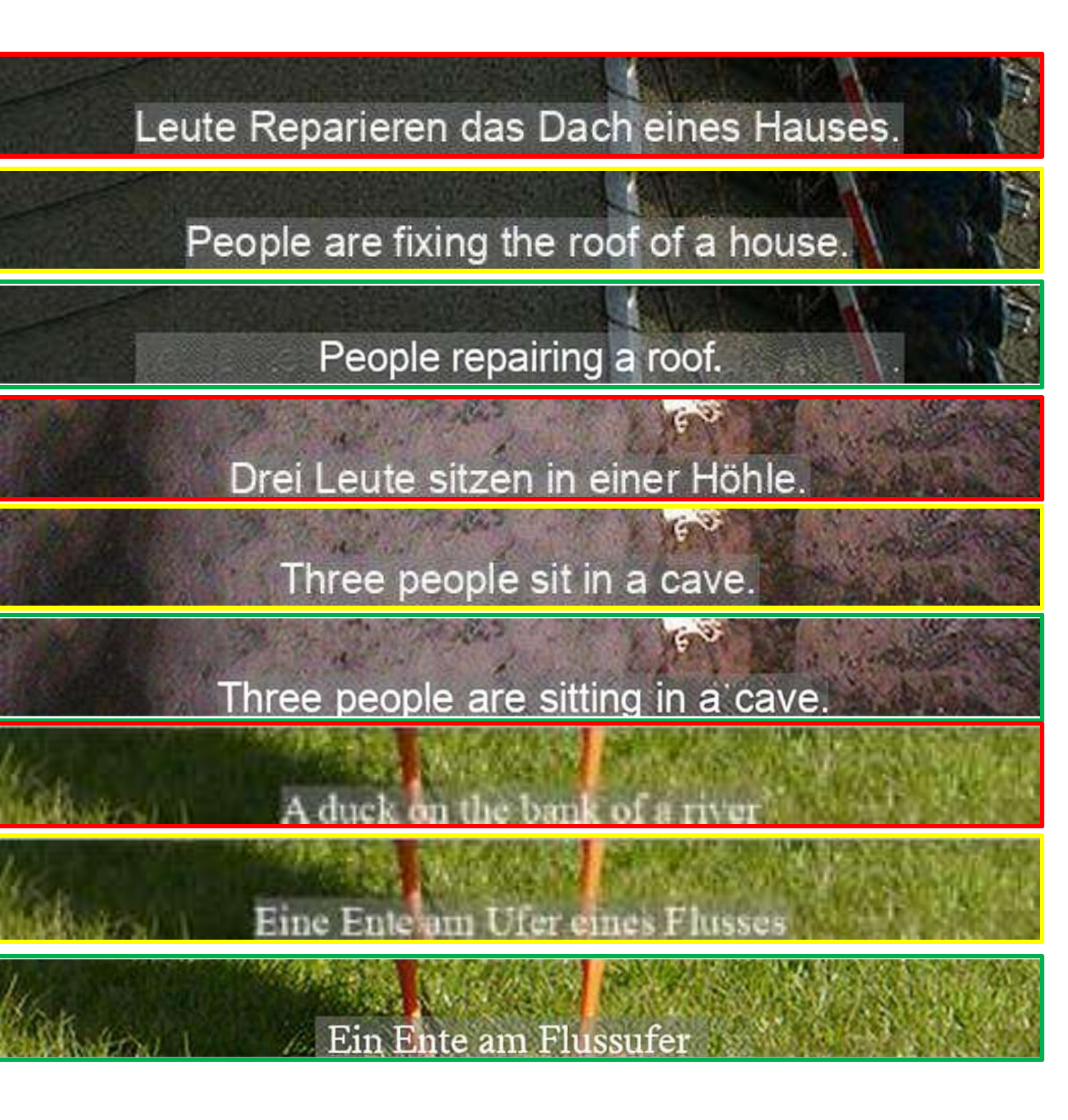}
    \caption{
Qualitative comparison on the IIMT30k
benchmark~\cite{tian2025exploring}. Images with red borders denote the source images, images with yellow borders correspond to the ground-truth targets provided in the original dataset, and images with green borders represent the outputs generated by our UniTranslator.
}
    \label{fig:vis_iimt30k}
\end{figure}

\begin{figure}[ht]
\centering
\includegraphics[width=0.7\textwidth]{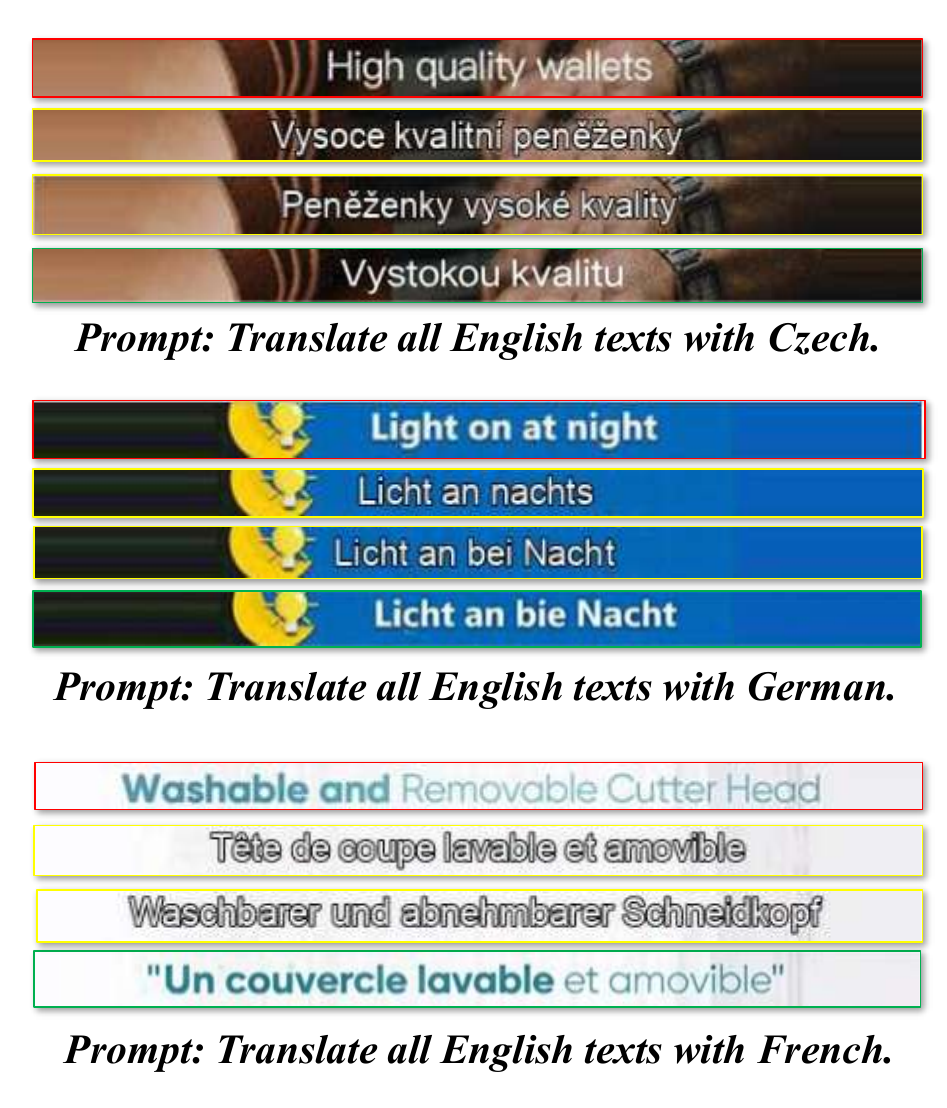}
    \caption{
Qualitative comparison on the PRIM benchmark~\cite{tian2025prim}. Red-bordered images denote the source images, green-bordered images represent the outputs generated by our UniTranslator, and yellow-bordered images correspond to the ground-truth targets provided in the dataset. The two ground-truth images are produced by first translating the source text using GPT-4 and Google Translate, respectively, and then rendering the translated text with a text-to-image (T2I) tool. And images with green borders represent the outputs generated by our UniTranslator.
}
    \label{fig:vis_prim}
\end{figure}

\cref{fig:vis_zs_anytrans} presents qualitative comparisons on the AnyTrans benchmark under zero-shot settings. The examples mainly consist of real-world scene texts. According to qualitative results, the UniTranslator generates translations that are both semantically accurate and visually well-aligned with the original scene. In particular, it preserves the background texture, color tone, and typography style of the source image while replacing the original text with fluent target-language translations, demonstrating its strong generalization ability for real-world scene text translation without additional task-specific training.

\cref{fig:vis_iimt30k} shows qualitative comparisons on the IIMT30k benchmark for IIMT. Beyond ensuring high translation quality and accurate text editing in German $\leftrightarrow$ English scenarios, UniTranslator also demonstrates strong font preservation ability across different styles. Specifically, it not only generates semantically faithful and fluent translations but also maintains the original typographic attributes of the source text, such as font shape, stroke thickness, and overall visual style. These results indicate that UniTranslator better balances translation correctness, layout alignment, and font-aware visual rendering in image-guided machine translation.

\cref{fig:vis_prim} presents qualitative results on the PRIM benchmark, where the task requires translating embedded English texts in product images into different target languages according to the given prompt. Compared with the ground-truth images rendered by a generic T2I tool in the original dataset, UniTranslator better preserves the visual characteristics of the original text, including font identity, color, stroke style, and overall typographic appearance, while seamlessly integrating the translated content into the image. This indicates that our method is more faithful to the source text style rather than generating a re-rendered typography that may deviate from the original design. Meanwhile, we observe some limitations in low-resource language settings, where UniTranslator occasionally produces missing words or inaccurate character rendering. We leave these issues for future work and aim to further improve translation completeness and fine-grained glyph generation quality in subsequent iterations.